\documentclass{bvm} 

\begin{document}

\newcommand{\bvmyear}{2025}

\selectlanguage{english} 

\title{Efficient Deep Learning-based Forward Solvers for Brain Tumor Growth Models}

\subtitle{}

\titlerunning{DL PDE Forward Solvers}

\author{
	Zeineb \lname{Haouari} \inst{1,2},
	Jonas \lname{Weidner} \inst{1,2},
    Yeray \lname{Martin-Ruisanchez}\inst{1},
    Ivan \lname{Ezhov} \inst{1,2},
    Aswathi \lname{Varma} \inst{1,2},
	Daniel \lname{Rueckert} \inst{1,2},
    Bjoern \lname{Menze} \inst{3},
	Benedikt \lname{Wiestler} \inst{1,2} 
}

\authorrunning{Haouari et al.} 

\institute{
\inst{1} Technical University of Munich\\
\inst{2} Munich Center for Machine Learning (MCML)\\
\inst{3} University of Zurich
}

\email{zeineb.haouari@tum.de}

\maketitle

\begin{abstract}
Glioblastoma, a highly aggressive brain tumor, poses major challenges due to its poor prognosis and high morbidity rates. Partial differential equation-based models offer promising potential to enhance therapeutic outcomes by simulating patient-specific tumor behavior for improved radiotherapy planning. However, model calibration remains a bottleneck due to the high computational demands of optimization methods like Monte Carlo sampling and evolutionary algorithms.
To address this, we recently introduced an approach leveraging a neural forward solver with gradient-based optimization to significantly reduce calibration time. This approach requires a highly accurate and fully differentiable forward model. We investigate multiple architectures, including (i) an enhanced TumorSurrogate, (ii) a modified nnU-Net, and (iii) a 3D Vision Transformer (ViT). The nnU-Net achieved the best overall results, excelling in both tumor outline matching and voxel-level prediction of tumor cell concentration. It yielded a low MSE ($2.7 \cdot 10^{-4}$) in tumor cell concentration compared to ground truth numerical simulation and the highest Dice score across all tumor cell concentration thresholds\footnote{In this revised version, we identified and corrected an implementation error in the nnU-Net, which has now led to improved performance}. Our study demonstrates significant enhancement in forward solver performance and outlines important future research directions.

Our source code is openly available at \url{https://github.com/ZeinebZH/TumorNetSolvers}
\end{abstract}

\section{Introduction}
Glioblastoma, the most prevalent malignant primary brain tumor, continues to have a dismal prognosis with a median overall survival of less than two years, despite extensive clinical trial efforts \cite{weller2021eano}. A significant challenge in improving treatment outcomes lies in the tumor's invasive nature, as it diffusely infiltrates surrounding brain tissue, rendering surgical resection of the entire tumor spread highly challenging. Consequently, postoperative radiotherapy (RT) has become a cornerstone of treatment, aimed at eliminating residual tumor cells while minimizing damage to healthy brain tissue. However, conventional imaging fails to detect these invading tumor cells, and RT planning relies on a uniform 15mm margin around the visible tumor to account for the diffuse spread, lacking patient-specific customization and limiting its effectiveness in addressing individual variations in tumor behavior \cite{niyazi2023estro}.

Tumor modeling based on partial differential equations (PDEs) has emerged as a promising tool for providing personalized insights into tumor growth and invasion dynamics, driving the development of more targeted and effective treatment strategies. High-precision modeling techniques such as Markov Chain Monte Carlo have demonstrated potential in personalizing tumor simulations \cite{lipkova2019personalized}. However, these methods involve numerous forward simulations, leading to extended runtimes, ranging from several hours to days, limiting their clinical applicability.

In contrast, deep learning (DL) methods present a faster alternative for solving this inverse problem by predicting tumor growth parameters without the need for iterative computations, thereby reducing runtimes to mere minutes \cite{ezhov2023learn, pati2021estimating}. Despite their speed, DL models face significant challenges in generalizing to unseen cases, which restricts their broader applicability, particularly in critical clinical settings such as RT planning. 

To overcome both these limitations, we recently proposed a novel idea that integrates the computational efficiency of DL models with the adaptability of gradient-based optimization \cite{weidner2024rapid}. Employing the DL-based forward solver TumorSurrogate \cite{ezhov2021geometry}, we conducted 
gradient-based optimization with respect to the input parameters of the tumor growth model. This iterative process minimizes the loss between the predicted and actual observed tumor distribution in the patient's MRI scans. In a cohort of nine glioma patients, our method significantly reduced the time required to solve the inverse problem from hours to minutes while achieving comparable modeling results.

Central to our novel gradient-based optimization approach is the DL-based forward solver. Here, we, therefore, systematically investigate distinct network architectures for their suitability as forward solvers. Our \emph{key contributions} in this work are:

\begin{itemize}
\item Repurposing the nnU-Net, originally designed for segmentation, to a conditioned regression-based tumor growth simulation, demonstrating best performance for this task.
\item Extending a vision transformer to 3D tumor growth prediction with explicit conditioning on anatomical context and tumor model parameters.
\item Systematically comparing these architectures with a baseline TumorSurrogate regarding tumor outline matching and voxel-level tumor cell concentration prediction.
\end{itemize}

\section{Materials and methods}

\subsection{Physical model}

Within the reaction-diffusion formalism, brain tumor growth is modeled as a combination of two primary processes: proliferation and diffusion. This dynamic is governed by a partial differential equation (PDE) that describes changes in the normalized tumor cell density \( c \) over time: 
\[
\frac{\partial c}{\partial t} = \nabla \cdot (D \nabla c) + \rho c (1 - c),
\]

where \( \nabla \cdot (D \nabla c) \) represents the spatial diffusion of tumor cells, and \( c(1-c) \) models logistic proliferation. The key parameters include the diffusion coefficient \( D \), the proliferation rate \( \rho \), and the tumor origin \( (x, y, z) \), collectively forming the parameter set \( \theta = \{\rho, D, x, y, z\} \). Periodic boundary conditions are applied to ensure computational stability. Patient-specific anatomical information is integrated through segmentation masks of white matter (WM), gray matter (GM), and cerebrospinal fluid (CSF). Tumor diffusion is restricted to the WM and GM regions, with \( D \) assigned a fixed ratio between these tissues and set to zero in the CSF regions, following the framework outlined in \cite{ezhov2021geometry}.

\subsection{Model architectures}

\emph{TumorSurrogate (TS):}
The baseline model, TumorSurrogate \cite{ezhov2021geometry}, employs an encoder-decoder convolutional architecture with residual skip connections and bottleneck parameter integration. The encoder downsamples brain tissue data through convolutional blocks to generate a latent representation, which is then conditioned at the bottleneck by projecting and concatenating a biophysical parameter vector. The decoder subsequently maps this combined representation to a tumor simulation. Skip connections are employed symmetrically within both the encoder and decoder.

\emph{Regression nnU-Net:}
We adapted nnU-Net \cite{isensee_nnu-net_2021} for conditioned image regression by modifying it to handle continuous data. Key changes include adapting the preprocessing pipeline (Subsection 2.3), replacing the combined loss function (Dice + Cross-Entropy loss) with MSE loss, and adapting the dynamic U-Net architecture. Specifically, we removed the softmax activation layer to enable continuous value prediction and conditioned the network on the input biophysical parameter vector, in the exact same manner as the baseline. Unlike TumorSurrogate, this architecture features U-Net-style residual skip connections between the encoder and decoder, and is fully automatically configured based on dataset properties.

\emph{Vision Transformer (ViT):}
We adapted the image reconstruction ViT from Lin and Heckel \cite{Lin2022VisionTE} for 3D conditioned image-to-image regression by extending it to 3D and incorporating biophysical parameter conditioning. Volumetric, non-overlapping patches ($16^3$) were encoded as spatial tokens, and a biophysical parameter vector was incorporated as a distinct token within the shared embedding space. This combined representation enables the model to jointly leverage spatial structure and tumor-specific information for accurate brain tumor simulation. The model comprises 12 transformer blocks (6 attention heads, embedding dimension 384, MLP ratio 4).

\begin{figure}[b]
		\includegraphics[width=0.7\figwidth]{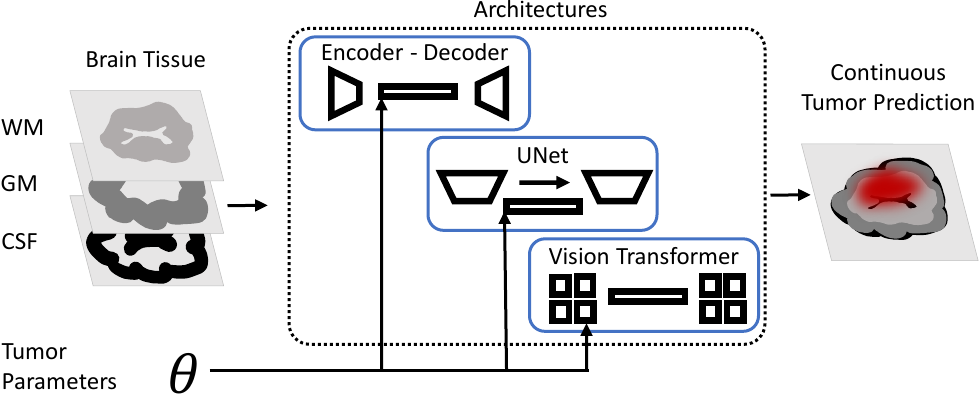}

		\caption{Overview of our experimental pipeline. We input brain anatomy maps and tumor parameters \( \theta \) into different networks (blue) and output the 3D tumor concentration (red).}
	\label{3675-overview} 
\end{figure}

\subsection{Training setup and experiments}
\emph{Dataset:} Models were trained and evaluated on a synthetically generated dataset (numerical solver \cite{balcerak2024physics}, brain tissue atlas geometry \cite{yeh2022population}) comprising 10,000 training, 2,000 validation, and 2,000 test samples. Inputs consisted of tissue data paired with corresponding biophysical parameters, with tumor simulations serving as targets.
\begin{table*}[t]
    \centering
    \caption{Training configurations for all models.}
    \label{3675-training_configurations}
    \begin{tabular*}{\textwidth}{l@{\extracolsep{0.2cm}}l@{\extracolsep{0.2cm}}l@{\extracolsep{0.2cm}}l}
        \hline
        Pipeline & Preprocessing & Loss Function & Other Details \\
        \hline
        ViT & Both pipelines & MSE & - \\
         & Data augmentation &  &  \\
        nnU-Net & Both pipelines & MSE, Deep supervision & - \\
         & Data augmentation &  &  \\
        TS Baseline & Original preprocessing & Region-specific MSE \cite{weidner2024rapid} & - \\
        TS Optimized & Both pipelines & MSE & Batch norm, Gradient \\
         & Data augmentation &  & clipping, Kaiming weight  \\
         & &  & initialization \\
        \hline
    \end{tabular*}
\end{table*}

\emph{Preprocessing:}
The original pipeline involved shifting the center of mass of the tumor to the image center, cropping to \(120 \times 120 \times 120\), and downsampling to \(64 \times 64 \times 64\) to reduce memory costs. We also adapted the nnU-Net preprocessing pipeline, retaining regression-suitable steps like normalization, resampling, and transposing while excluding segmentation-specific steps like foreground and class-based sampling.

\emph{Training Configurations:} The nnU-Net model follows the predefined configuration in \cite{isensee_nnu-net_2021}, employing Stochastic Gradient Descent (SGD) with a momentum coefficient of 0.99 and Nesterov acceleration. A polynomial decay schedule manages the learning rate, starting from $1 \times 10^{-2}$. The TS Model is trained with the Adam optimizer, using a weight decay of $4 \times 10^{-20}$ and parameters $\beta = (0.9, 0.999)$. A cosine annealing scheduler varies the learning rate between $10^{-6}$ and $10^{-4}$ as suggested in \cite{ezhov2021geometry}. The ViT model is trained with the AdamW optimizer, with a learning rate range of $2 \times 10^{-4}$ to $8 \times 10^{-4}$ and a weight decay of $1 \times 10^{-2}$. Default $\beta$ parameters $(0.9, 0.999)$ are used, and a one-cycle learning rate scheduler with a 10\% ramp-up period is applied.

\emph{Data augmentation}
follows the nnU-Net framework, discarding intensity-based augmentations as they are unsuitable for tissue data, keeping rotations and mirroring.

This results in four different pipelines (Tab. \ref{3675-training_configurations})

\section{Results}
\subsection{Voxel-level prediction of tumor cell concentration}
In Table \ref{3675-metrics_summary}, we compare voxel-level accuracy in predicting tumor cell concentration across different models (measured by MAE and MSE) as well as the global image similarity (SSIM) on the unseen test set. Notably, the nnU-Net model consistently outperforms all other models across all metrics, achieving an MSE that is a quarter of that of the second-best model.
In addition, we observe that our optimized TumorSurrogate outperforms the baseline model by a factor of 2 in terms of the MSE.

\begin{table}[t]
    \centering
    \caption{Summary of model performances. We report mean and standard error.}
    \label{3675-metrics_summary}
    \begin{tabular*}{\textwidth}{l@{\extracolsep\fill}ccc}
        \hline
        Model &  MSE [$ 10^{-3}$] ($\downarrow$) &  MAE [$ 10^{-3}$] ($\downarrow$) &  SSIM ($\uparrow$) \\
        \hline
        ViT & 3.27 $\pm$ 0.06  & 15.57 $\pm$ 0.22  & 0.878 $\pm$ 0.001 \\
        nnU-Net  & 0.27 $\pm$ 0.03 & 3.64 $\pm$ 0.18 & 0.987 $\pm$ 0.001 \\
        TS Baseline & 2.02 $\pm$ 0.04  & 12.27 $\pm$ 0.16  & 0.903 $\pm$ 0.002 \\
        TS optimized & 0.98 $\pm$ 0.02  & 8.21 $\pm$ 0.09 & 0.912 $\pm$ 0.001 \\
        \hline
    \end{tabular*}
\end{table}

\subsection{Tumor outline matching}
\begin{figure}[b]
		\includegraphics[width=0.8\figwidth]{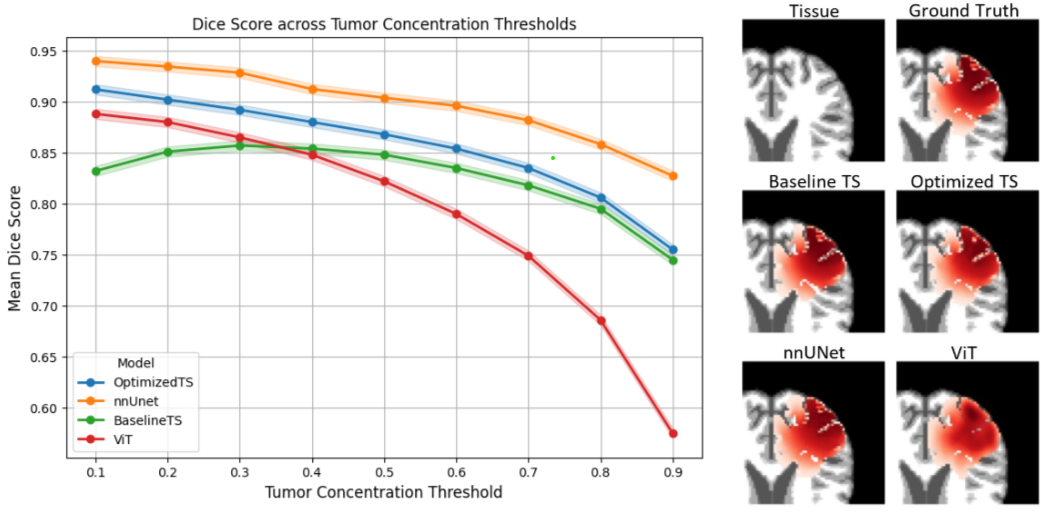}

		\caption{(left) Comparison of Dice overlap for the different network architectures across varying tumor concentration thresholds. We plot the mean Dice score and standard error on the test set. (right) Example images of tumor concentration prediction.}
	\label{3675-dice} 
\end{figure}

To effectively apply the solver for planning radiotherapy in tumor patients, where only visible tumor margins are available for model fitting, it is crucial to assess how accurately the tumor outline is matched by the different models. In Figure \ref{3675-dice}, we investigate the Dice overlap with the ground truth tumor cell distribution across various tumor cell concentration $c$ thresholds. Again, the nnU-Net model performs best overall. Notably, at low tumor cell concentration thresholds (up to $c \sim 0.3$), the ViT model performs almost on par with the optimized TumorSurrogate. However, its performance declines significantly at higher thresholds, translating to smaller tumor volumes, which may be attributed to the coarse tokenization employed (with patches sized 16×16×16).

\section{Discussion}
Our recently introduced gradient-based approach to solving the inverse problem \cite{weidner2024rapid} relies on a reliable, differentiable forward solver. This study explores three deep learning architectures: an encoder-decoder model (TumorSurrogate), a nnU-Net-based model, and a ViT-based model. The nnU-Net outperforms all other methods in predicting voxel-level tumor cell concentration and matching tumor outlines. Thus, the implicit bias and empirical hyperparameter tuning of the nnU-Net translate well to this uniform grid-based simulation.
Moreover, we observe that the optimized TumorSurrogate model outperforms the baseline in terms of matching the tumor outline, especially in lower tumor cell concentrations.
For clinical translation, matching the visible tumor outline (Dice score) is crucial. The ViT model showed promise at lower tumor cell concentrations but performed worse at higher thresholds, likely due to its coarse patch-based tokenization. Future work will improve tokenization and explore conditional diffusion models.
A crucial next step is translating these findings to real patient data and extending to more complex tumor growth models, including mass effect or necrosis. While this work uses solely a synthetic dataset, the focus must shift to validating the developed models on real patient datasets. We propose training the models on synthetic data derived from diffusion model-generated MRI segmentations as simulation geometries, followed by fine-tuning on a fraction of real patient data. This transfer learning approach leverages the extensive knowledge gained from synthetic data while adapting to real patient characteristics.

\printbibliography

\end{document}